\documentclass[10pt,twocolumn,letterpaper]{article}

\pdfoutput=1

\newcommand{\eref}[1]{Eq.~(\ref{#1})}

\newcommand{\fref}[1]{Fig.~\ref{#1}}
\newcommand{\Fref}[1]{Figure~\ref{#1}}
\newcommand{\sref}[1]{Sec.~\ref{#1}}

\usepackage{iccv}
\usepackage{times}
\usepackage{epsfig}
\usepackage{graphicx}
\usepackage{amsmath}
\usepackage{amssymb}
\usepackage{enumitem}
\usepackage{color}
\usepackage{subfigure}
\usepackage{multirow}
\usepackage[percent]{overpic}
\usepackage{xcolor}

\usepackage[pagebackref=true,breaklinks=true,letterpaper=true,colorlinks,bookmarks=false]{hyperref}

\iccvfinalcopy 

\def\iccvPaperID{869} 

\ificcvfinal\fi
\begin{document}

\title{Modelling the Scene Dependent Imaging in Cameras \\ 
with a Deep Neural Network}

\author{Seonghyeon Nam\\
Yonsei University\\
{\tt\small shnnam@yonsei.ac.kr}
\and
Seon Joo Kim\\
Yonsei University\\
{\tt\small seonjookim@yonsei.ac.kr}
}

\maketitle

\begin{abstract}
We present a novel deep learning framework that models the scene dependent image processing inside cameras.
Often called as the radiometric calibration, the process of recovering RAW images from processed images (JPEG format in the sRGB color space) is essential for many computer vision tasks that rely on physically accurate radiance values.
All previous works rely on the deterministic imaging model where the color transformation stays the same regardless of the scene and thus they can only be applied for images taken under the manual mode.
In this paper, we propose a data-driven approach to learn the scene dependent and locally varying image processing inside cameras under the automode.
Our method incorporates both the global and the local scene context into pixel-wise features via multi-scale pyramid of learnable histogram layers.
The results show that we can model the imaging pipeline of different cameras that operate under the automode accurately in both directions (from RAW to sRGB, from sRGB to RAW) and we show how we can apply our method to improve the performance of image deblurring. 
\end{abstract}

\section{Introduction}
Deep learning has significantly changed the approaches for solving computer vision problems.
Instead of analytic solutions with some combinations of hand chosen features, probabilistic/physical models and some optimizations, most methods now turn to deep learning which is a deeper neural network that relies on big data.
Deep learning has shown superb performance in many computer vision problems including image recognition~\cite{he2016deep}, face recognition~\cite{schroff2015facenet}, segmentation~\cite{long2015fully}, etc. 
Image processing problems such as super-resolution~\cite{Dong14,Kim16} and colorization~\cite{Larsson16,Zhang16} are also solved with deep learning now, which provides effective ways to process input images and output images that fit the given task. 

In this paper, we introduce a new application of using deep learning for image processing: modelling the scene dependent image processing. 
We are especially interested in modelling the in-camera imaging pipeline to recover RAW images from camera processed images (usually in the form of JPEG in the sRGB color space) and vice versa.
Usually called as the radiometric calibration, this process is important for many computer vision tasks that require accurate measurement of the scene radiance such as photometric stereo~\cite{Jung_2015_CVPR}, intrinsic imaging~\cite{bell14intrinsic}, high dynamic range imaging~\cite{Debevec}, and hyperspectral imaging~\cite{Oh2016}. 

There are two strategies with regards to the image processing in cameras, namely, the photographic reproduction model and the 
photofinishing model~\cite{Holm}. 
In the photographic reproduction model, the image rendering pipeline is fixed meaning that a raw RGB value will always be mapped to an RGB value in the processed image regardless of the scene. 
Taking photos under the manual mode triggers this model and all previous radiometric calibration methods work only in this mode. 

\begin{figure}
\center
\subfigure[Manual mode]{\includegraphics[width=0.47\linewidth]{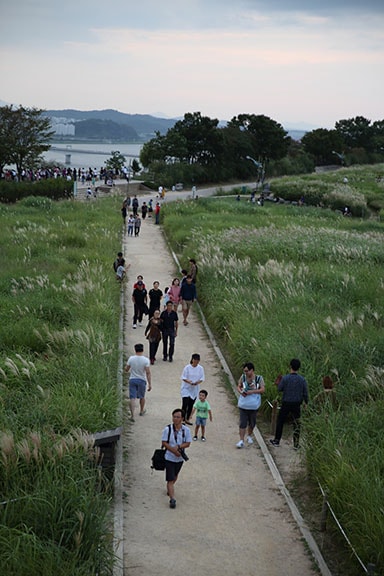}}
\subfigure[Auto-mode]{\includegraphics[width=0.47\linewidth]{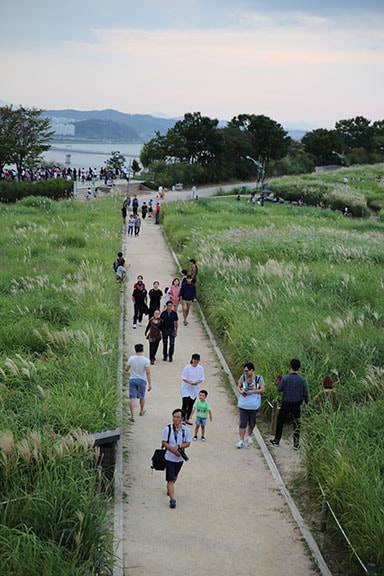}}
\caption{Difference of two images captured (a) under the manual mode and (b) under the auto-mode. The RAW images of both (a) and (b) are identical. In (b), the brightness/contrast and the colors were enhanced automatically by the camera.}
\label{fig:SceneDependency}
\end{figure}

In the photofinishing model, the image processing inside the camera varies (possibly in a spatially varying manner) in order to deliver visually optimal picture depending on the shooting environment~\cite{CanonBrochure}.
This scene dependent mode will be activated usually when the camera is operated under the auto-mode.
Figure~\ref{fig:SceneDependency} compares the photos of a scene captured under the manual mode and under the auto-mode. 
In (b), the scene became brighter and the colors were enhanced compared to (a).
It shows that the color rendering will be dependent on the scene in the auto-mode.
Scene dependency in cameras were also verified in \cite{Chakrabarti}.
As mentioned above, none of the previous work can deal with the scene dependent color rendering.
This is a problem as there are many photometry related topics in computer vision that have access to only automatically taken images (e.g. internet images) as in \cite{bell14intrinsic,Jung_2015_CVPR}. 
Moreover, smartphone cameras have become the major source for images and the many phone cameras only work in the auto-mode.  

The goal of this paper is to present a new algorithm that can model the camera imaging process under the "auto" mode.  
To deal with the scene dependency, we take the data-driven approach and design a deep neural network.
We show that modelling the image processing using conventional CNN-based approaches is not effective for the given task, 
and propose a multi-scale pyramid of learnable histogram~\cite{Wang16} to incorporate both the global and the local color histogram into pixel-wise features.
The extracted multi-scale contextual features are processed with our CNN to model the scene dependent and locally varying color transformation.

To the best of our knowledge, this is the first paper that can extract RAW images from processed images taken under the auto setting.
Being able to radiometrically calibrate smartphone cameras is especially a significant contribution of this work. 
We show that we can model both the forward rendering (RAW to sRGB) and the reverse rendering (sRGB to RAW) accurately using our deep learning framework. 

We further apply our work to image deblurring. 
A blurred image is first transformed to the RAW space, in which a deblurring algorithm is executed.
The deblurred image is then transformed back to the sRGB space through our deep network. 
We show that performing deblurring in this fashion give much better results over deblurring in the nonlinear sRGB space.

\begin{figure*}
\center
\subfigure[Canon EOS 5D Mark III]{\includegraphics[width=0.3\linewidth]{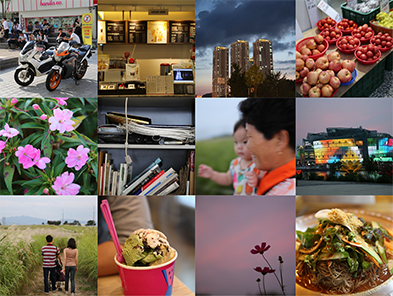}}\hspace{0.5cm}
\subfigure[Nikon D600]{\includegraphics[width=0.3\linewidth]{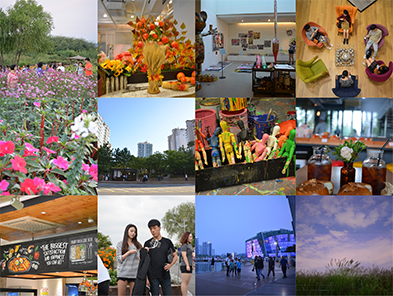}}\hspace{0.5cm}
\subfigure[Samsung Galaxy S7]{\includegraphics[width=0.3\linewidth]{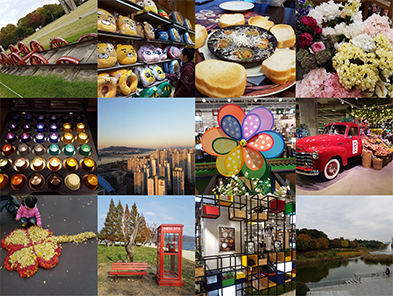}}
\caption{Examples of images in our dataset. The dataset covers a wide range of scenes and colors.}
\label{fig:Dataset}
\end{figure*}

\section{Related Work}
\noindent \textbf{In-camera Image Processing (Radiometric Calibration)} 
\\
In the early works of radiometric calibration, the relationship between the scene radiance and the image value was explained  just by a tonemapping function called the camera response function. 
Different models of the response function~\cite{Grossberg_PAMI_04,Mann,Mitsunaga} as well as robust estimation algorithms~\cite{Debevec,Kim_PAMI_08,Lin_CVPR04} were introduced.
More comprehensive reviews of the radiometric calibration literature is presented in \cite{Kim12}.

In \cite{Kim12}, a more complete in-camera imaging pipeline that includes processes such as the white balance, the color space transformation, and the gamut mapping in addition to the tone-mapping was introduced. 
With the new parametric model for the imaging, the work also presented an algorithm for recovering the parameters from a set of images using a scattered point interpolation scheme. 
Using a similar pipeline, a probabilistic model that takes into account the uncertainty in the color rendering was recently proposed in \cite{Chakrabarti2014}.

As mentioned earlier, all previous works are based on the assumption that the color rendering inside the camera is deterministic and therefore cannot be applied for photos taken under the automode or by phone cameras.
In comparison, our deep network framework learns the scene dependent image processing through given data and thus can be used for automatically captured photos.

\vspace{0.3cm}
\noindent \textbf{Deep Learning for Low-level Vision} 
\\
Deep learning has been very successful in image classification tasks, and the deep neural networks are now being applied to various problems in computer vision including the low-level vision tasks.
In the field of low-level vision, convolutional neural networks (CNNs) are used to exploit the local context information for various applications such as image super-resolution~\cite{Dong14,Kim16}, denoising~\cite{Jain09,NIPS16}, and filtering~\cite{Li16,Xu15}. While the input and the output of these applications are RGB images, the learned mapping is more of a structural mapping rather than being a color mapping.

Recently, deep learning based image colorization has been studied~\cite{Larsson16,Zhang16}, of which the objective is to restore chrominance information from a single channel luminance image. These works exploit the high-level semantic information to determine the chrominance of pixels by using CNNs, similar to those used in the high-level recognition tasks~\cite{Simonyan14}. 
In this paper, we show that color histogram based low-level features extracted using our deep network are more efficient for the given task compared to the high-level CNN features extracted from above previous work.

In~\cite{Yan16}, an automatic photo adjustment method using a multi-layer perceptron was proposed. 
They feed the concatenation of global color features and semantic maps to a neural network system to find the scene dependent and the locally varying color mapping. 
As with the other data-driven image enhancement techniques~\cite{Bychkovsky2011,Hwang2012}, the features for the color mapping in their work are manually selected. 
However, one of the key properties behind the success of deep learning is in its ability to extract good features for the given task automatically. 
Instead of using manual features, we propose an end-to-end trainable deep neural network to model the scene dependent in-camera image processing.

\section{Dataset}
An essential ingredient for any deep learning system is a good dataset.
To model the image processing inside the camera from data, we need pairs of RAW images and its corresponding images in the nonlinear sRGB color space with JPEG format. 
Using the RAW-JPEG shooting mode, which is now supported by most cameras including Android based smartphones, we can collect many pairs of corresponding RAW and sRGB images.
In this paper, we collected images using three digital cameras: Canon 5D Mark III, Nikon D600, and Samsung Galaxy S7.
All pictures were taken under the auto-mode and the features like Auto Lighting Optimizer in the Canon camera that triggers locally varying processing such as contrast enhancement were all turned on. 
Some of the images in our dataset are shown in \Fref{fig:Dataset}. 
As can be seen, our dataset contains various kind of scenes including outdoor, indoor, landscape, portrait, and colorful pictures.
The number of images in the dataset are 645, 710, and 290 for the Canon, the Nikon, and the Samsung camera, respectively.
50 images of varying scenes for each camera were selected for the test sets.
In training phase, we extract multiple patches from images on the fly by using the patch-wise training method, which is described in~\sref{sec:patch_training}.
Therefore, we can make millions of training examples from hundreds of image data.

One thing to take notice is the white balancing in the imaging pipeline.
The white balance is one of the main factors in determing the image color.
While the white balance factor can be learned in the forward pipeline (from RAW to JPEG) as shown in \cite{Oh2017}, estimating the white balance in the reverse pipeline is seemingly a more difficult task as the illumination is already normalized in the JPEG image.
After the images are illumination normalized with the white balancing, it becomes an one-to-many mapping problem as any illuminant could have been mapped to the current image.

Fortunately, the meta information embedded in images (EXIF data) provides the white balancing information.
It provides three coefficients, which are the scale factors for the red, the green, and the blue channels. 
All the RAW images in our dataset are first white balanced using this information from the EXIF data. 
Therefore, the mapping that we learn in our system is from the white balanced RAW to sRGB, and vice versa. 
%
%
\begin{figure*}
\center
\includegraphics[width=0.8\linewidth]{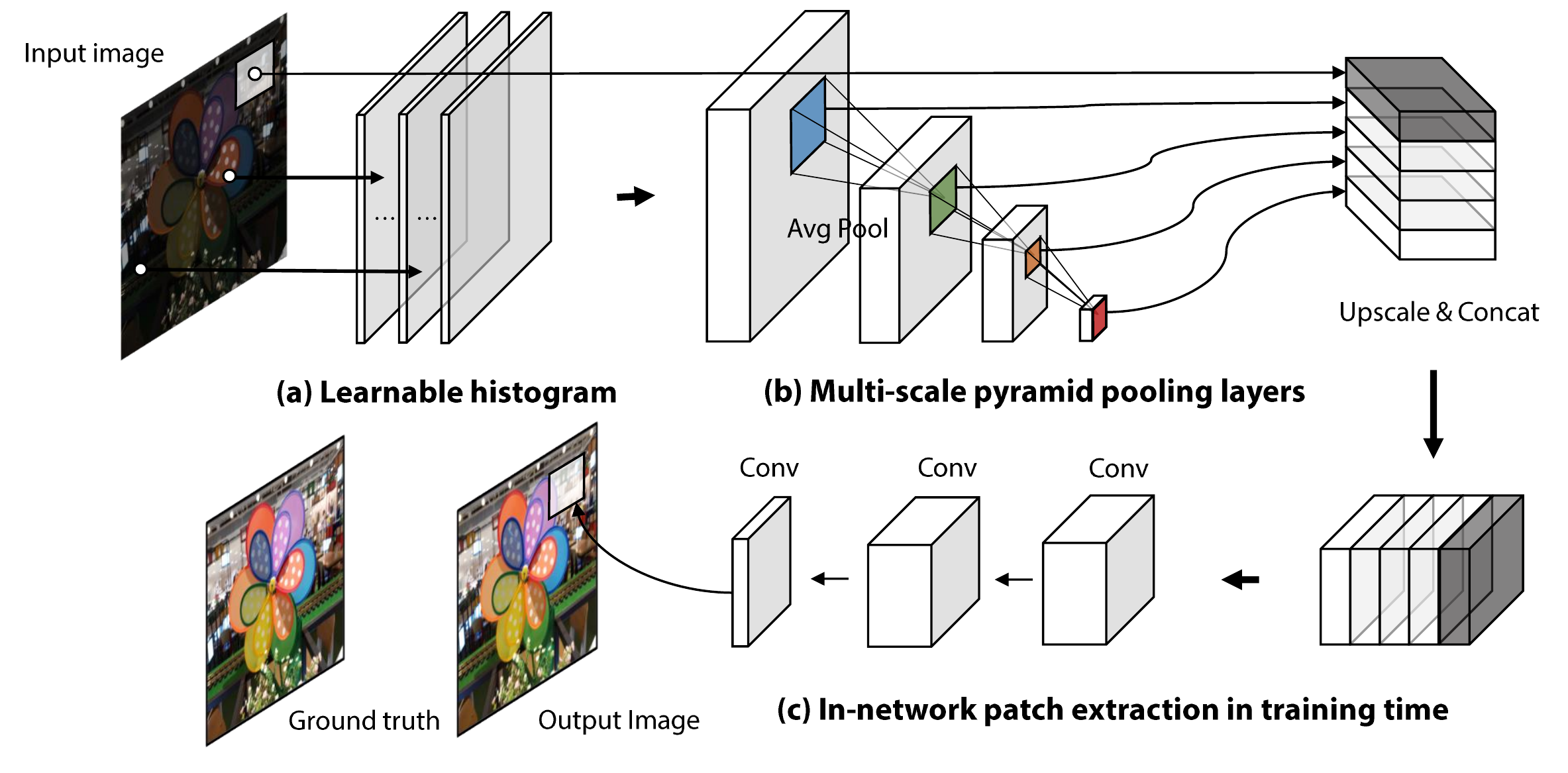}
\caption{The overview of the proposed deep neural network.}
\label{fig:overview}
\end{figure*}
\section{Deep Learning Framework for Modelling the Imaging Pipeline}
In this work, our goal is to model the imaging pipeline by computing the function $f$ that maps RAW images to sRGB images and the function $f^{-1}$ that maps sRGB images to RAW images. 
The training data consist of RAW-sRGB pairs $\mathcal{D}=\{X^i, Y^i\}_{i=0}^n$, where $X$ is the RAW image, $Y$ represents the sRGB image, and $n$ is the number of training examples.
Since the deep neural networks are not invertible, we train $f$ and $f^{-1}$ separately.
Without the loss of generality, the algorithm that follows will be explained for the forward mapping $f$.
Exactly the same process can be applied for learning the reverse mapping $f^{-1}$.

The mapping function $f$ under the auto-mode varies according to the scene and the local neighborhood.
The function is formally described as: 
\begin{equation}
Y^i_x = f(X^i_x, \Phi^i, \Omega^{i}_{x}),
\label{eq:model}
\end{equation}
where $i$ is the image index, $x$ is the pixel index, $\Phi$ represents the global scene descriptor, and $\Omega$ indicates local descriptor around a pixel. 
We propose a deep neural network that learns the scene dependent color mapping $f$ including both $\Phi$ and $\Omega$ in~\eref{eq:model} in an end-to-end manner.

To optimize the parameters of the proposed network, we need to define the loss function that computes the difference between the estimation and the ground truth. 
We minimize $l_2$ error from the training data as follows:
\begin{equation}
L = \frac{1}{n}\sum_{i=0}^{n}\Vert f(X^i)-Y^i \Vert^2.
\end{equation}

As explained, the color mapping $f$ is dependent on the global and the local context.
Coming up with features that can describe this scene dependency manually is a difficult task. 
One way to compute the features for this problem is to use pre-trained CNNs like the VGG network~\cite{Simonyan14} and finetune using our training set.
As we show in Section 5, applying conventional CNN based structures do not capture good features for the scene dependency in our task. 
From the camera's point of view, it would be difficult to run a high-level scene recognition module for the scene dependent rendering due to the computational load. 
Therefore, it is reasonable to conjecture that the scene dependent color mapping relies mostly on low-level features such as 
the contrast and the color distributions, which are computationally cheaper than the semantic features. In this work, we exploit color histogram as the feature to describe the scene.

\subsection{Learnable Histogram}
Color histogram is one of the most widely used features to describe images. In deep networks that use histograms, the centers and the widths of the bins are hand-tweaked by the user. In addition, since the computation of histograms is not differentiable, histograms are precomputed before training deep networks. Meanwhile, Wang \etal~\cite{Wang16} recently proposed the learnable histogram method, in which the key is a specialized differentiable function that trains the optimized histogram from data with deep networks in an end-to-end manner.

\begin{figure}
\center
\includegraphics[width=0.9\linewidth]{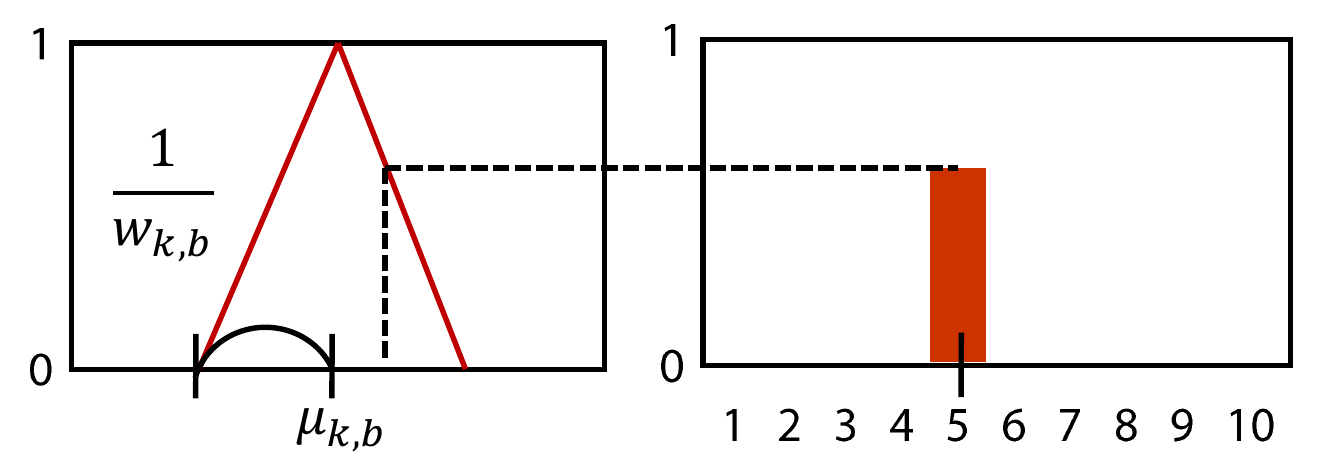}
\caption{The concept of histogram voting function of the learnable histogram~\cite{Wang16}. The left is the histogram voting function, and the right is an example of it.}
\vspace{-0.3cm}
\label{fig:learnable_histogram}
\end{figure}

With the learnable histogram, the bin for the value of an element in the feature map is determined by the following voting function:
\begin{equation}
\psi_{k,b}(x_k) = \text{max}\{ 0, 1-\vert x_k - \mu_{k,b} \vert \times w_{k,b} \}.
\label{eq:learnable_histogram}
\end{equation}
$k$ and $b$ are the index of the feature map element and the output bin, respectively.
$x_k$ is the value of the $k$-th element in the feature map, $\mu_{k,b}$ and $w_{k,b}$ are the center and the width of the $b$-th bin.  
The histogram is built by accumulating the bins computed with the function $\psi_{k,b}(x_k)$ as illustrated in \fref{fig:learnable_histogram}. 
The centers $\mu_{k,b}$ and the widths $w_{k,b}$ are trainable parameters and are optimized together with other parameters of the deep network.

In this work, we adopt the histogram voting function~\eref{eq:learnable_histogram} of the learnable histogram to extract image features for our task of modelling the imaging pipeline.
By introducing a multi-scale pyramid of histograms, we design the pixel-wise color descriptor for the global and the local context.
In~\cite{Wang16}, the learnable histogram was applied to the intermediate semantic feature maps to exploit global connect global Bag-of-Words descriptors, which in turn improves the performance of semantic segmentation and object detection.
As we are looking for more low level color features instead of high-level semantic features, we directly connect the learnable histogram to the input image to extract RGB color histograms as show in~\fref{fig:overview} (a).
Moreover, by putting multi-scale pooling layers on the output of the learnable histogram, our new network can extract the global and the local descriptors for each pixel.

To effectively extract the color distribution, it is necessary to decouple the brightness and the chromaticity distribution. 
Therefore, we first convert the RGB values to a lightness ($L$) and chromaticity ($rg$) channels before the image goes through the learnable histogram as follows:
\begin{equation}
\begin{split}
L &= (R + G + B) / 3,\\
r &= R / (R+G+B),\\
g &= G / (R+G+B).
\end{split}
\end{equation}

\subsection{Multi-Scale Pyramid Pooling Layer}
The output dimension of \eref{eq:learnable_histogram} is $H \times W \times (C \times B)$, where $H$, $W$ and $C$ are the number of the height, width and channel of the input, $C$ is the number of bins of the histogram. To get the global and the local color histogram, multiple average poolings with different pooling size are applied to the output feature maps as shown in \fref{fig:overview} (b). We concatenate the multi-scale features corresponding to the same input pixel to incorporate the global and local context into pixel-wise features. Formally, our multi-scale pyramid of histogram features is described as:
\begin{equation}
\Omega^{i}_{x} = [h_{x}^{1}, h_{x}^{2}, ..., h_{x}^{s}],
\end{equation}
where $h_{x}^{s}$ is the feature vector of $s$-th scale of the histogram layer corresponds to the pixel $x$.

In our implementation, we compute four scales of the multi-scale pyramid by cascading three $3\times3$ average pooling layers followed by a global average pooling for the global histogram.
The strides of the 3 local histogram layers are 1, 2, and 2, respectively.

\subsection{Patch-Wise Training Method and Implementation Details}
\label{sec:patch_training}
As illustrated in \fref{fig:overview}, our deep network is trained with image patches instead of using the whole image. 
In the training phase, the whole image is first forwarded to the learnable histogram module (\fref{fig:overview} (a)).
Then a number of patches are randomly selected from both the input image and the histogram feature maps (\fref{fig:overview} (b)).
Specifically, patches are first extracted from the input image, and the feature maps that correspond to each patch are cropped to form the multi-scale features.
Finally, only those selected patches are used for training the CNN weights as shown in \fref{fig:overview} (c).

This patchwise training has the advantage of being able to generate many training examples from a small dataset as well as being efficient in both time and memory. 
At the test time, the whole image and feature maps are forwarded to generate the full size output.

For the configuration of our network, we used 6 bins for the learnable histogram, the initial bin centers were set to (0, 0.2, 0.4, 0.6, 0.8, 1.0), and the initial bin widths were set to 0.2 as described in \cite{Wang16}.
After the global and the local features are extracted using the learnable histogram, the descriptors are concatenated with the input RGB image. 
Then, we apply $1\times1$ convolution filters to mix all input pixels and feature information, followed by two $3\times3$ additional convolutions to estimate the output.

\section{Experiments}
\label{sec:experiments}
\subsection{Experimentation setup}
The training images are preprocessed as follows. 
The RAW images are first demosaiced, normalized to have the max value of 1, and white-balanced using the EXIF metadata.
Images are downsized and cropped to 512$\times$512 images.
In all camera datasets, we use 80\% of images for the training and the remaining 20\% for the validation, excluding the 50 test images.
For the training, we use the Adam optimizer~\cite{kingma2014adam} to minimize our cost function. 
The batch size is 4, and sixteen $32\times32$ sparse patches are randomly extracted from it, which makes 64 training examples per batch.
According to~\cite{BansalChen16}, our training with a small fraction of images does not affect the convergence.
With a GTX 1080 GPU, we can train the proposed network of 100 epochs within an hour.

As explained before, we cannot compare the proposed method with existing radiometric calibration methods as they are deterministic models for specific manual settings and cannot be applied to automode cameras.
Instead, we compare the proposed method with the following four baseline methods.
\begin{description}
	\item[$\bullet$ Multi-layer Perceptron:] 
		We designed a MLP that consists of two hidden layers with 64 nodes each.
		The MLP learns an RGB to RGB color mapping without considering the scene dependency.
		We implemented the MLP by applying $1 \times 1$ convolution to images.
	\item[$\bullet$ SRCNN~\cite{Dong14}:] 
		We used the SRCNN that consists of five $3 \times 3$ convolutional layers without pooling, and this is a simple attempt to model the scene dependency.
	\item[$\bullet$ FCN~\cite{long2015fully} and HCN~\cite{BansalChen16,Hariharan15,Larsson16}:] 
		Since we only have hundreds of images in the training data, we adopt a pixel-wise sampling method~\cite{BansalChen16,Larsson16} to a hypercolumn network (HCN) to generate sufficient training signals. It cannot be applied to FCN since sample position is usually a fractional number in downsampled feature maps.
		Note that we only use VGG network layers from (\texttt{conv1\_1} to \texttt{conv4\_3}) for the FCN and (\texttt{conv1\_2}, \texttt{conv2\_2}, \texttt{conv3\_3}) for the HCN, since our machine cannot handle large feature maps computed from high-definition input images (e.g. 1920$\times$2880). For the FCN, we use the FCN-8S configuration on the reduced VGG network. We finetune both the FCN and the HCN using the pretrained VGG network.
\end{description}

\begin{table*} [!]
\setlength{\tabcolsep}{2.0mm}
\begin{center}
{\footnotesize
\begin{tabular}{|c|c|c|c|c|c|c|c|c|c|c|c|c|c|}
\hline
Rendering & \multirow{2}{*}{Methods} & \multicolumn{4}{|c|}{Canon 5D Mark III} & \multicolumn{4}{|c|}{Nikon D600} & \multicolumn{4}{|c|}{Samsung Galaxy S7} \\
\cline{3-14}
Setting & & Mean & Median & Min & Max & Mean & Median & Min & Max & Mean & Median & Min & Max \\
\hline \hline
\multirow{5}{*}{RAW-to-sRGB} & MLP & 25.43 & 25.58 & 18.08 & 32.46 & 27.63 & 27.84 & \textbf{22.95} & 30.93 & 27.62 & 27.90 & 24.96 & 30.56 \\
\cline{2-14}
& SRCNN~\cite{Dong14} & 27.62 & 27.52 & 18.25 & 35.38 & 27.90 & 27.92 & 22.81 & 32.17 & 30.03 & 30.04 & \textbf{26.21} & \textbf{34.40} \\
\cline{2-14}
& FCN~\cite{long2015fully} & 20.18 & 20.40 & 12.97 & 24.30 & 20.70 & 20.98 & 17.05 & 23.34 & 20.80 & 20.84 & 16.27 & 26.87 \\
\cline{2-14}
& HCN~\cite{Larsson16} & 27.53 & 28.04 & 18.84 & 34.78 & 27.99 & 28.30 & 22.73 & 33.20 & 29.05 & 29.27 & 26.02 & 32.79 \\
\cline{2-14}
& Ours & \textbf{29.63} & \textbf{29.94} & \textbf{19.24} & \textbf{36.32} & \textbf{28.85} & \textbf{28.93} & 22.41 & \textbf{33.86} & \textbf{30.14} & \textbf{30.63} & 22.03 & 33.91 \\
\hline \hline
\multirow{5}{*}{sRGB-to-RAW} & MLP & 34.72 & 34.77 & 26.35 & \textbf{44.19} & 32.77 & 32.04 & 24.38 & 40.77 & 29.56 & 29.99 & 23.23 & 34.28 \\
\cline{2-14}
& SRCNN~\cite{Dong14} & 32.34 & 32.98 & 21.30 & 39.43 & 30.51 & 29.57 & 23.00 & 38.05 & 30.12 & 31.44 & 22.45 & 35.35 \\
\cline{2-14}
& FCN~\cite{long2015fully} & 21.46 & 21.02 & 17.18 & 28.95 & 20.58 & 20.44 & 15.67 & 25.17 & 21.05 & 21.08 & 18.06 & 26.05 \\
\cline{2-14}
& HCN~\cite{Larsson16} & 33.49 & 33.46 & 25.16 & 39.70 & 32.99 & 32.39 & 26.63 & 39.68 & 30.02 & 30.76 & 22.95 & 34.86 \\
\cline{2-14}
& Ours & \textbf{35.16} & \textbf{35.38} & \textbf{26.73} & 42.80 & \textbf{33.67} & \textbf{33.35} & \textbf{27.61} & \textbf{42.02} & \textbf{31.67} & \textbf{32.66} & \textbf{25.10} & \textbf{37.39} \\
\hline
\end{tabular}
}
\vspace{1.5mm}
\caption{Quantitative result. The values are 4 statistics (mean median, min, max) of PSNRs in 50 test images. Bold text indicates the best performance.}
\label{table:quantitative}
\vspace{-3.5mm}
\end{center}
\end{table*}

\begin{figure*}
\centering
\begin{tabular}{@{}c@{ }c@{ }c@{ }c@{ }c@{ }c@{}}
\textsc{\small{Ground Truth}} & \textsc{\small{Ours}} & \textsc{\small{SRCNN~\cite{Dong14}}} & \textsc{\small{Error (ours)}} & \textsc{\small{Error (SRCNN~\cite{Dong14})}} \\
\vspace{-0.5mm}
\includegraphics[width=0.18\linewidth]{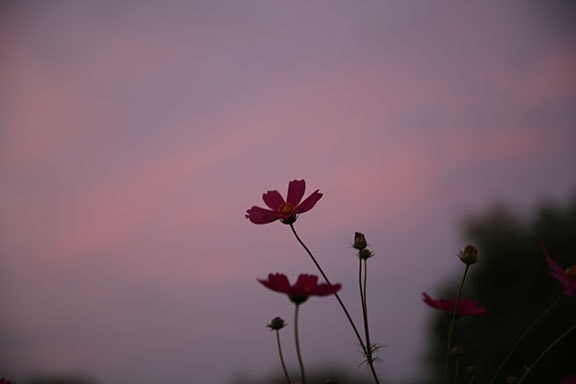}
&\includegraphics[width=0.18\linewidth]{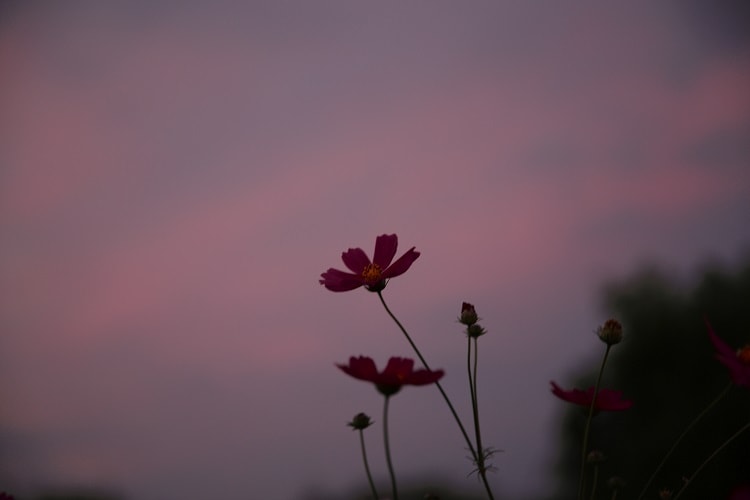}
&\includegraphics[width=0.18\linewidth]{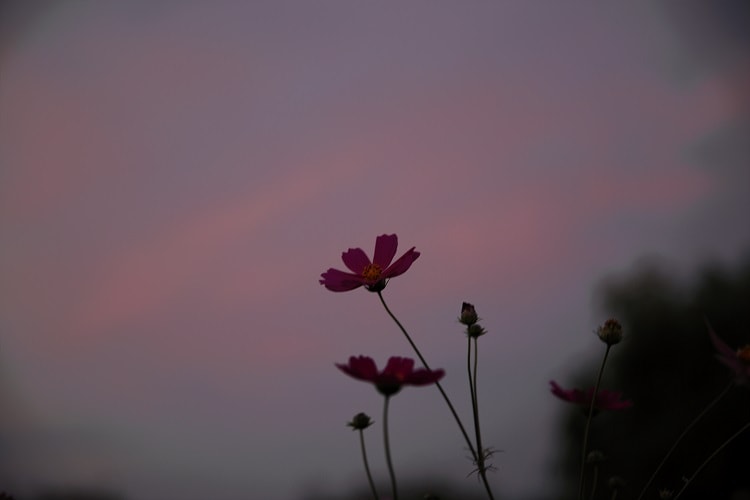}
&\includegraphics[width=0.18\linewidth]{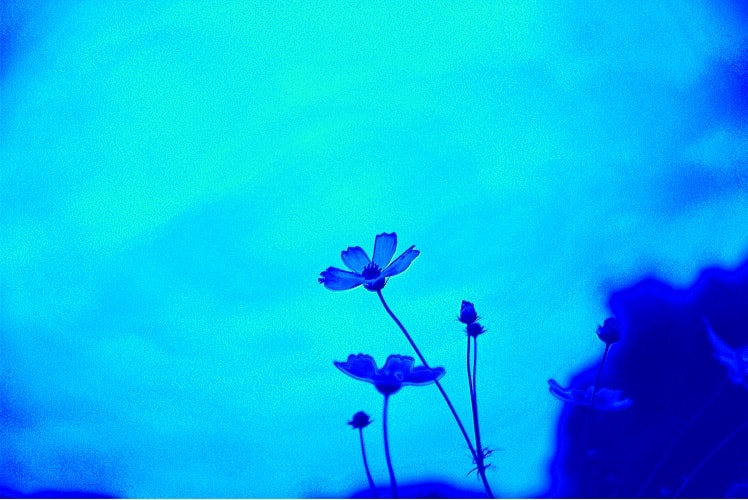}
&\includegraphics[width=0.18\linewidth]{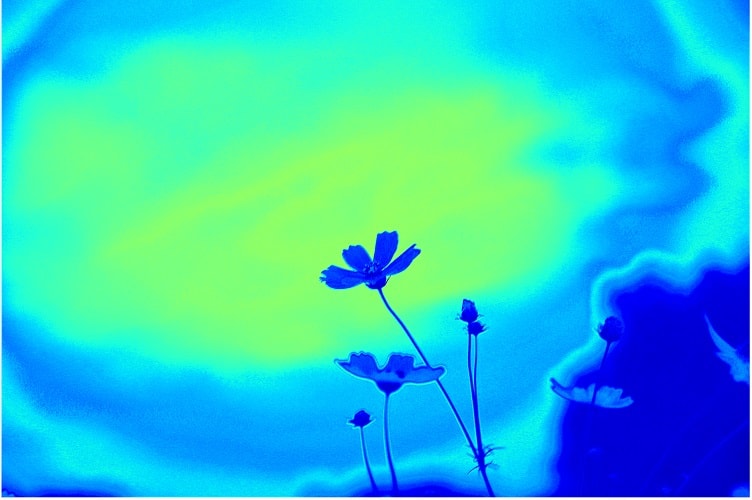}
&\includegraphics[width=0.035\linewidth]{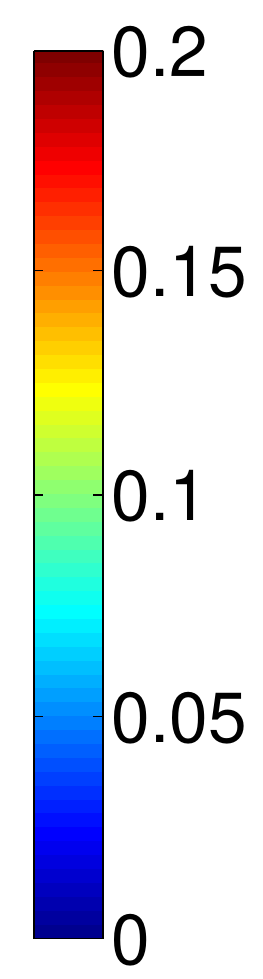}
\end{tabular}
\vspace{0.8mm}
\begin{tabular}{@{}c@{ }c@{ }c@{ }c@{ }c@{ }c@{}}
\includegraphics[width=0.18\linewidth]{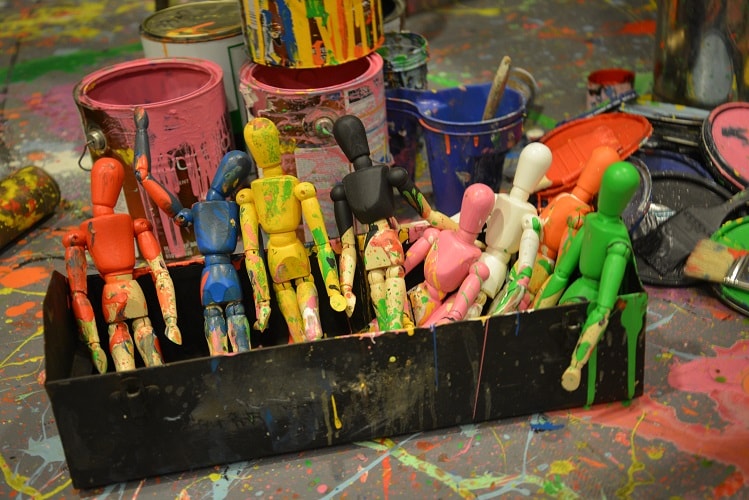}
&\includegraphics[width=0.18\linewidth]{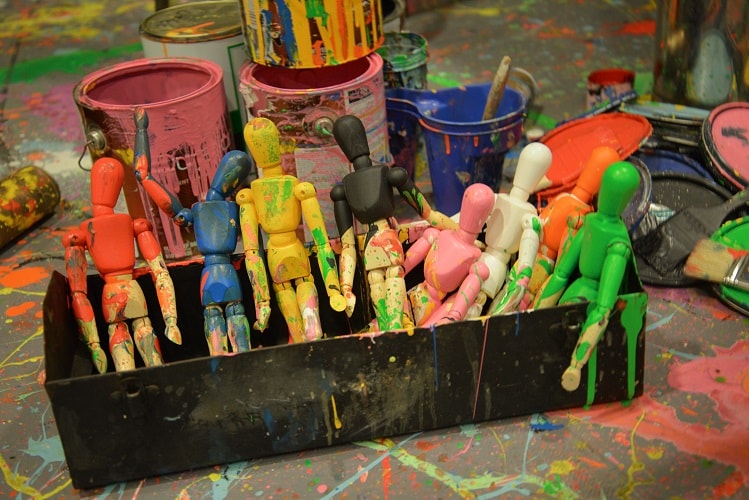}
&\includegraphics[width=0.18\linewidth]{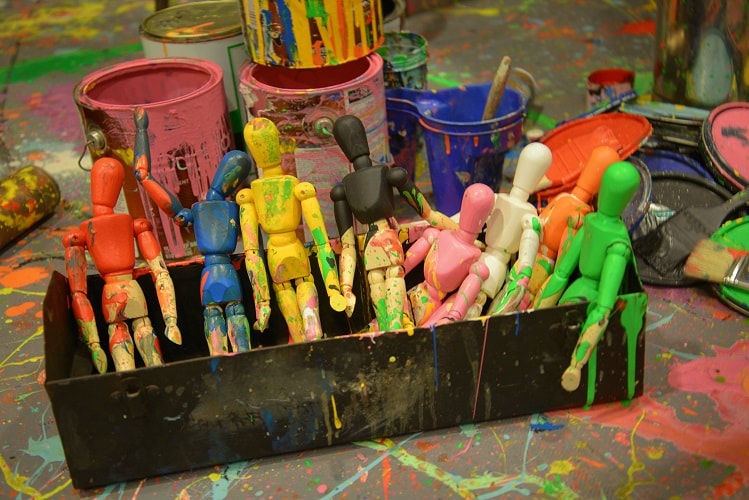}
&\includegraphics[width=0.18\linewidth]{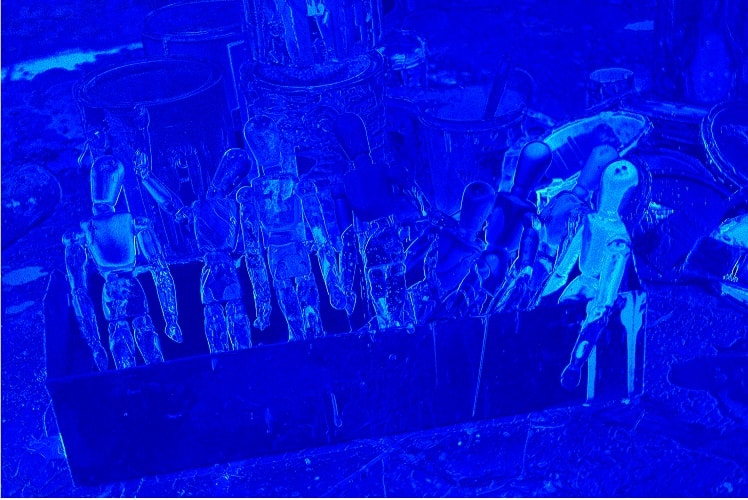}
&\includegraphics[width=0.18\linewidth]{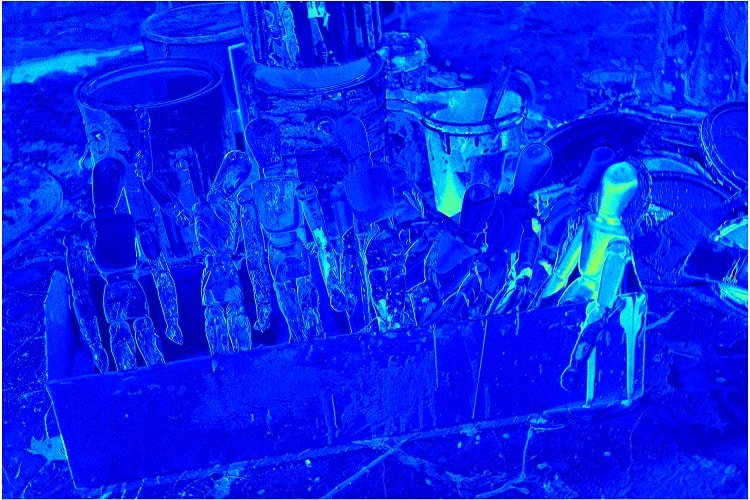}
&\includegraphics[width=0.035\linewidth]{./figure/results/colorbar_02}
\end{tabular}
\vspace{0.8mm}
\begin{tabular}{@{}c@{ }c@{ }c@{ }c@{ }c@{ }c@{}}
\includegraphics[width=0.18\linewidth]{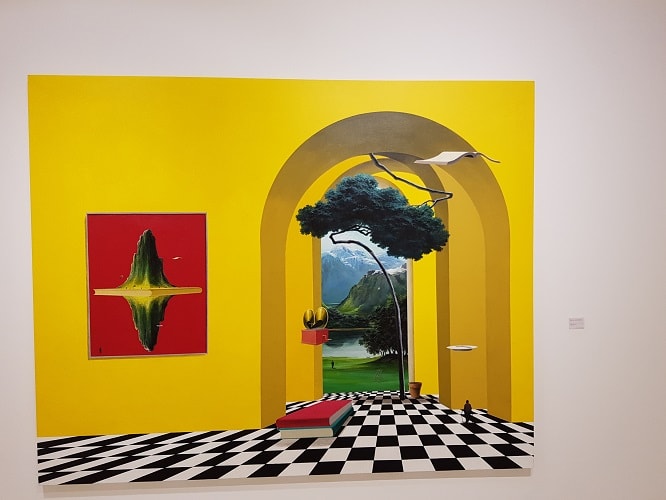}
&\includegraphics[width=0.18\linewidth]{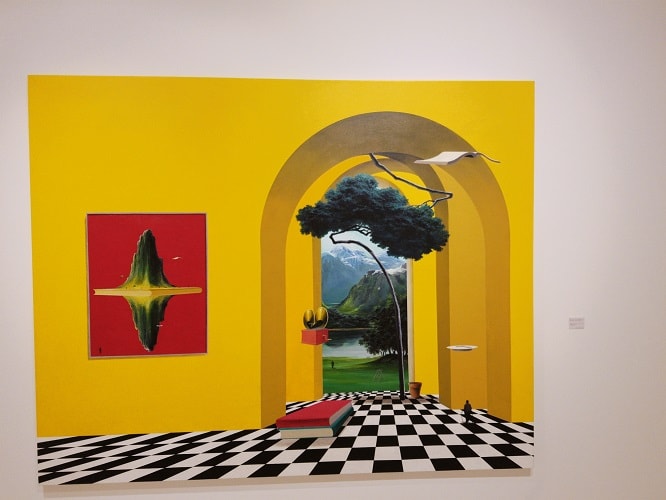}
&\includegraphics[width=0.18\linewidth]{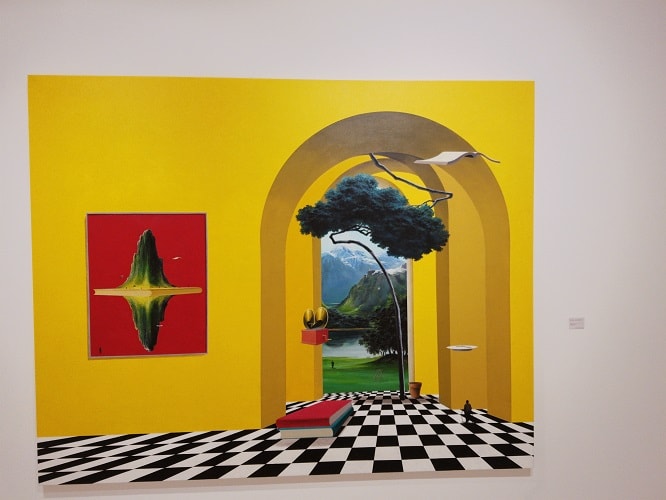}
&\includegraphics[width=0.18\linewidth]{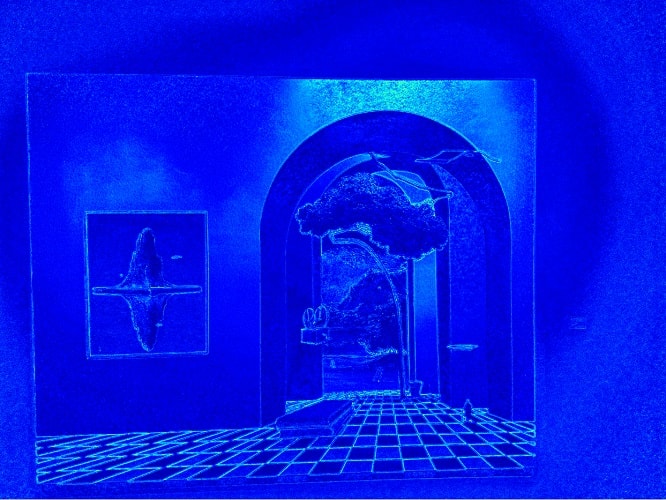}
&\includegraphics[width=0.18\linewidth]{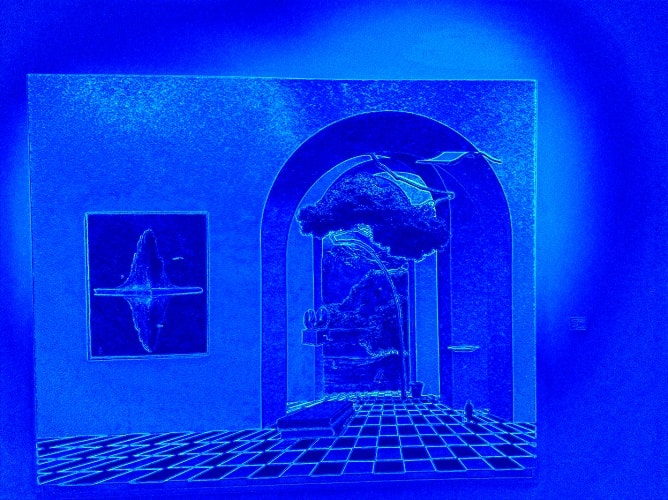}
&\includegraphics[width=0.04\linewidth]{./figure/results/colorbar_02}
\end{tabular}

\vspace{1.5mm}

\begin{tabular}{@{}c@{ }c@{ }c@{ }c@{ }c@{ }c@{}}
\includegraphics[width=0.18\linewidth]{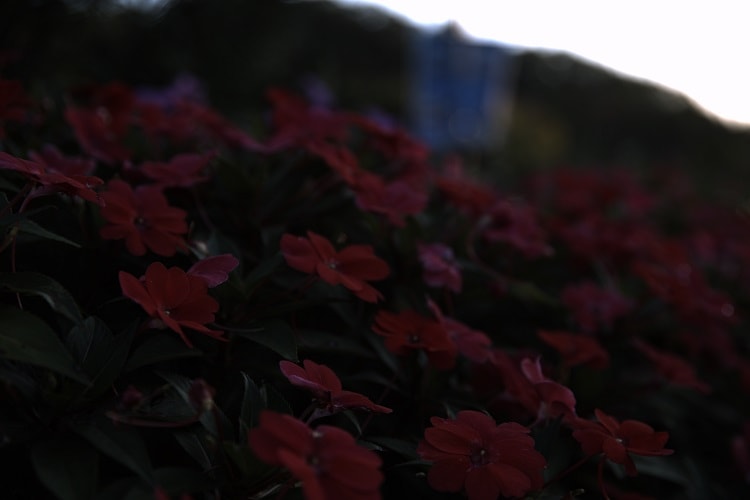}
&\includegraphics[width=0.18\linewidth]{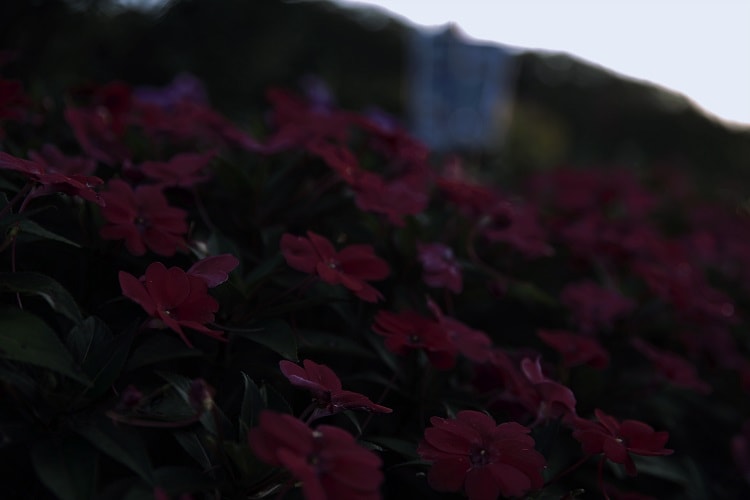}
&\includegraphics[width=0.18\linewidth]{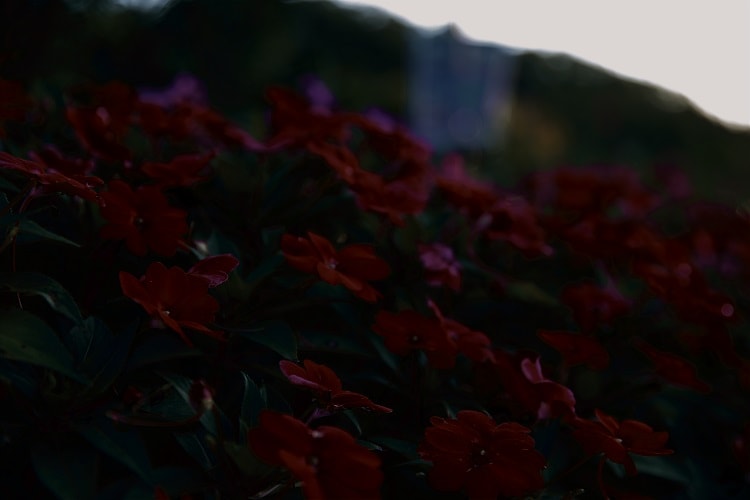}
&\includegraphics[width=0.18\linewidth]{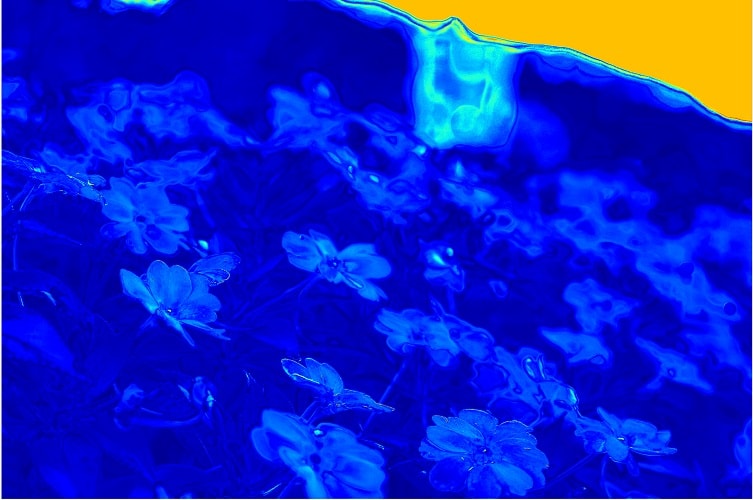}
&\includegraphics[width=0.18\linewidth]{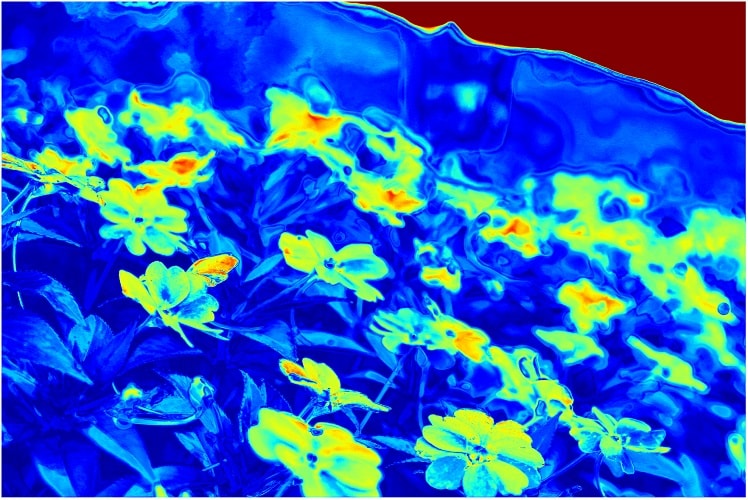}
&\includegraphics[width=0.04\linewidth]{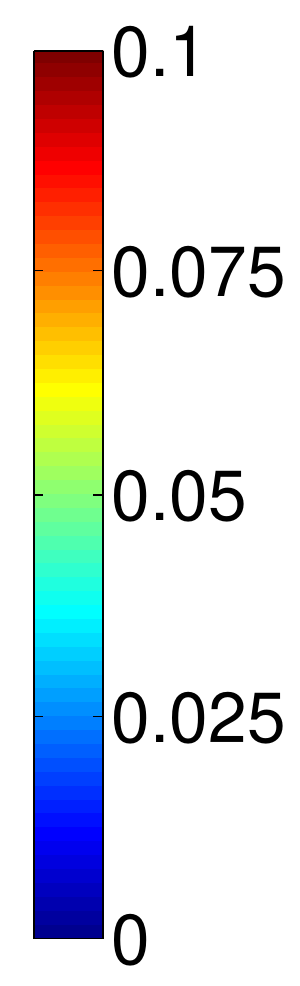}
\end{tabular}
\vspace{0.8mm}
\begin{tabular}{@{}c@{ }c@{ }c@{ }c@{ }c@{ }c@{}}
\includegraphics[width=0.18\linewidth]{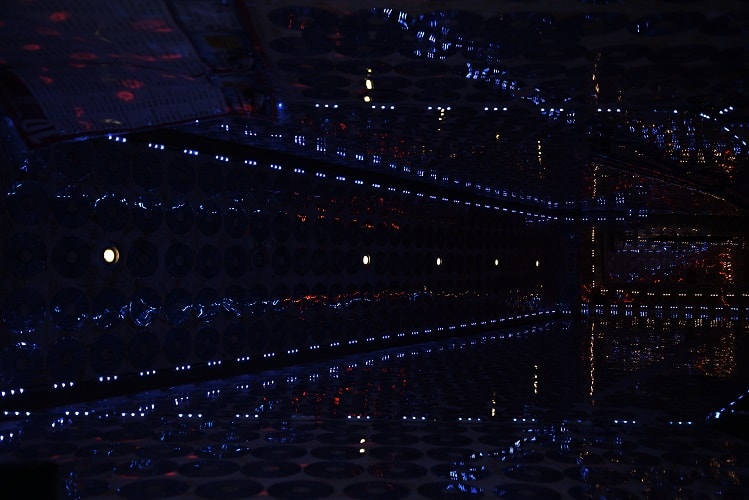}
&\includegraphics[width=0.18\linewidth]{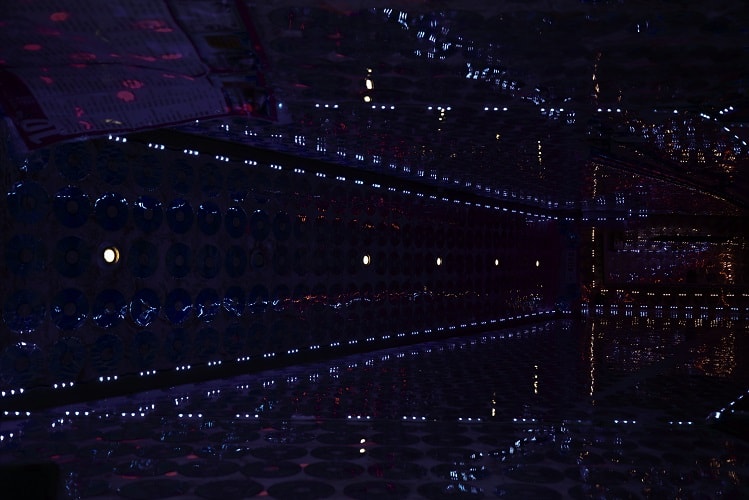}
&\includegraphics[width=0.18\linewidth]{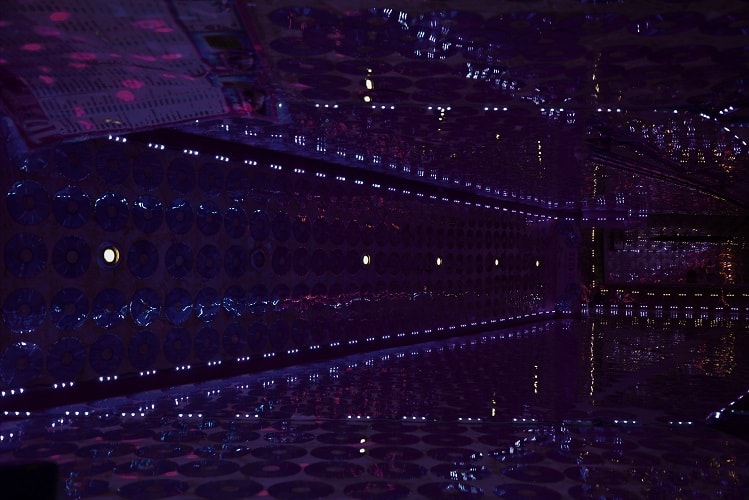}
&\includegraphics[width=0.18\linewidth]{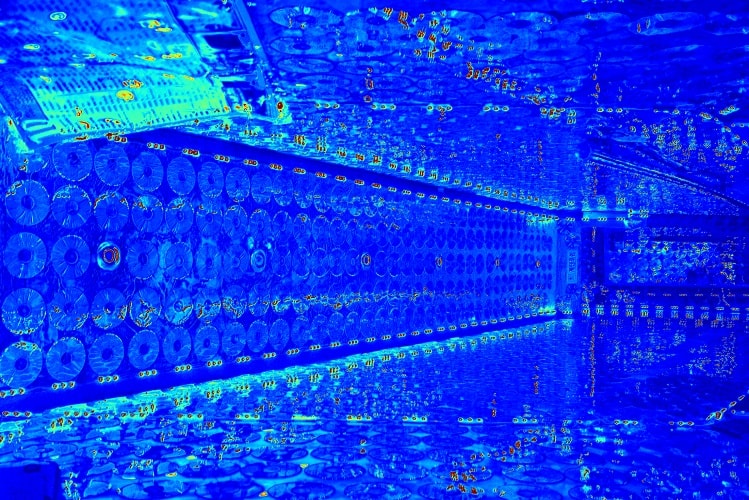}
&\includegraphics[width=0.18\linewidth]{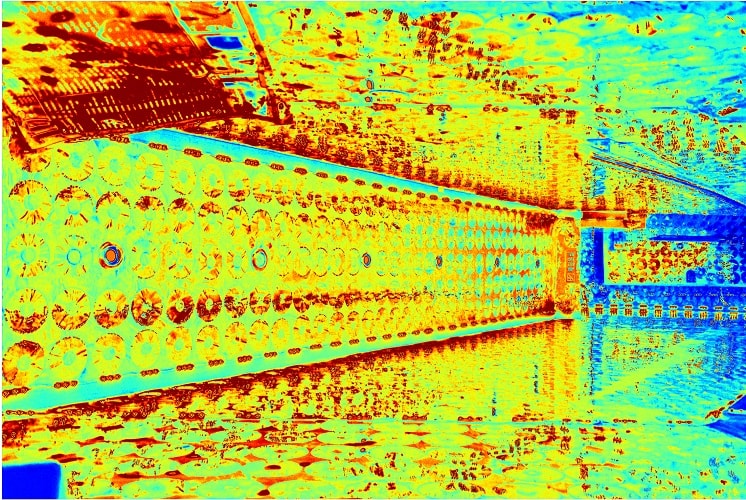}
&\includegraphics[width=0.04\linewidth]{./figure/results/colorbar_01}
\end{tabular}
\vspace{0.8mm}
\begin{tabular}{@{}c@{ }c@{ }c@{ }c@{ }c@{ }c@{}}
\includegraphics[width=0.18\linewidth]{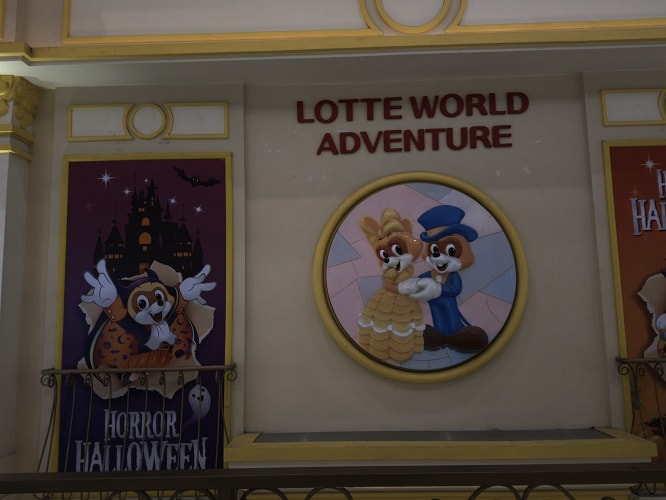}
&\includegraphics[width=0.18\linewidth]{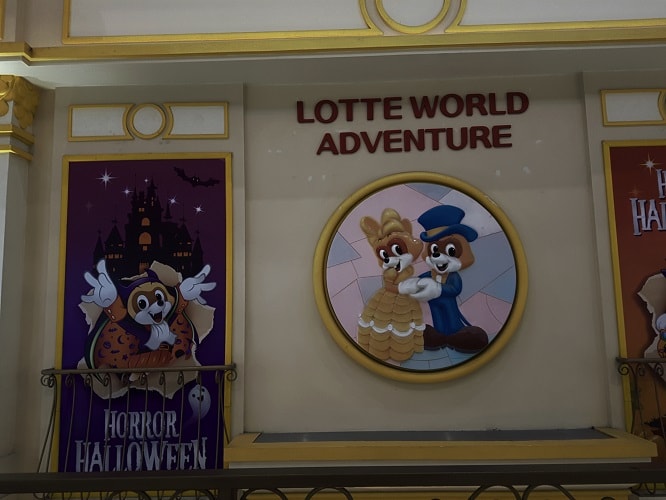}
&\includegraphics[width=0.18\linewidth]{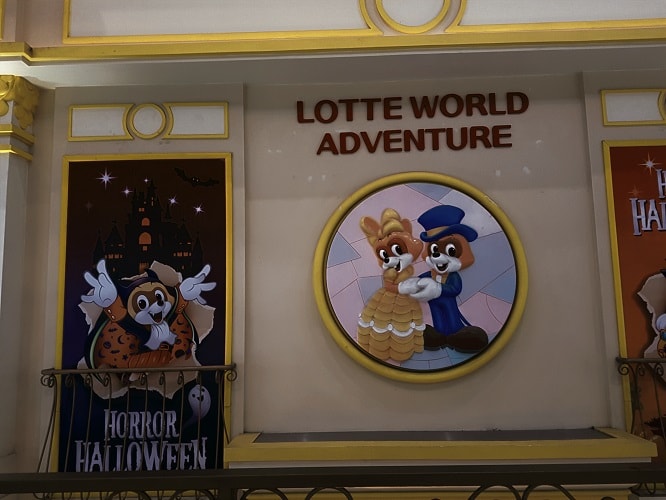}
&\includegraphics[width=0.18\linewidth]{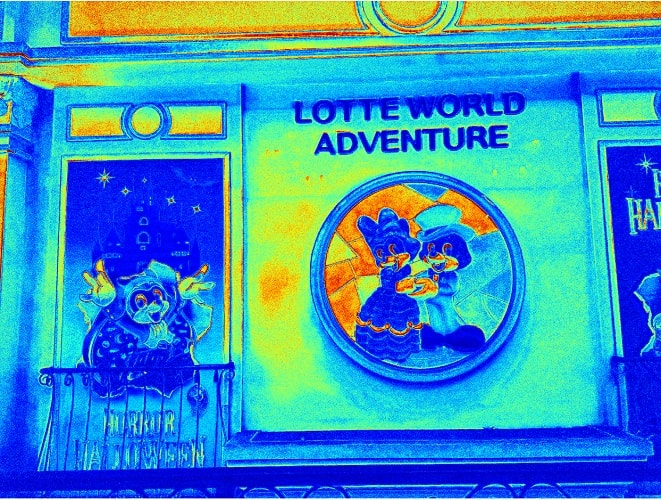}
&\includegraphics[width=0.18\linewidth]{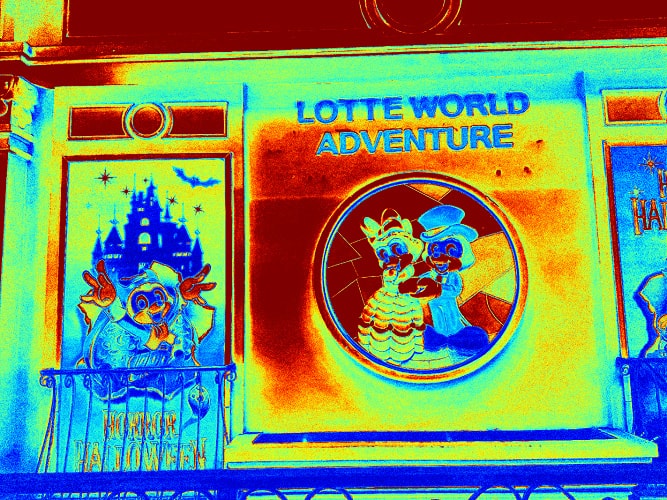}
&\includegraphics[width=0.045\linewidth]{./figure/results/colorbar_01}
\end{tabular}

\caption{Qualitative comparisons of results. The top 3 rows are the RAW-to-sRGB results of Canon 5D Mark III, Nikon D800, and Samsung Galaxy S7, and the bottom rows are the inverse mapping results of them, respectively.}
\vspace{1cm}
\label{fig:qualitative}
\vspace{-15.5mm}
\end{figure*}

\begin{figure*}
\center
\subfigure[The histogram of RAW]
{
\begin{overpic}[width=0.23\linewidth]{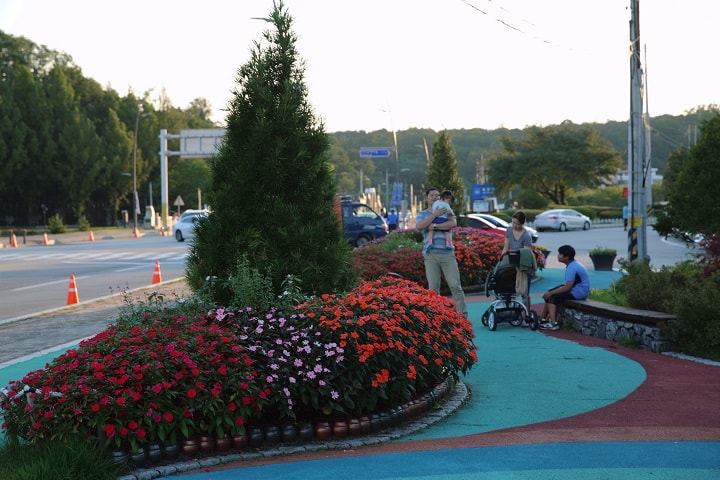}
\put(46, 1.8){\fbox{\includegraphics[trim=2px 2px 2px 2px, clip, scale=0.1]{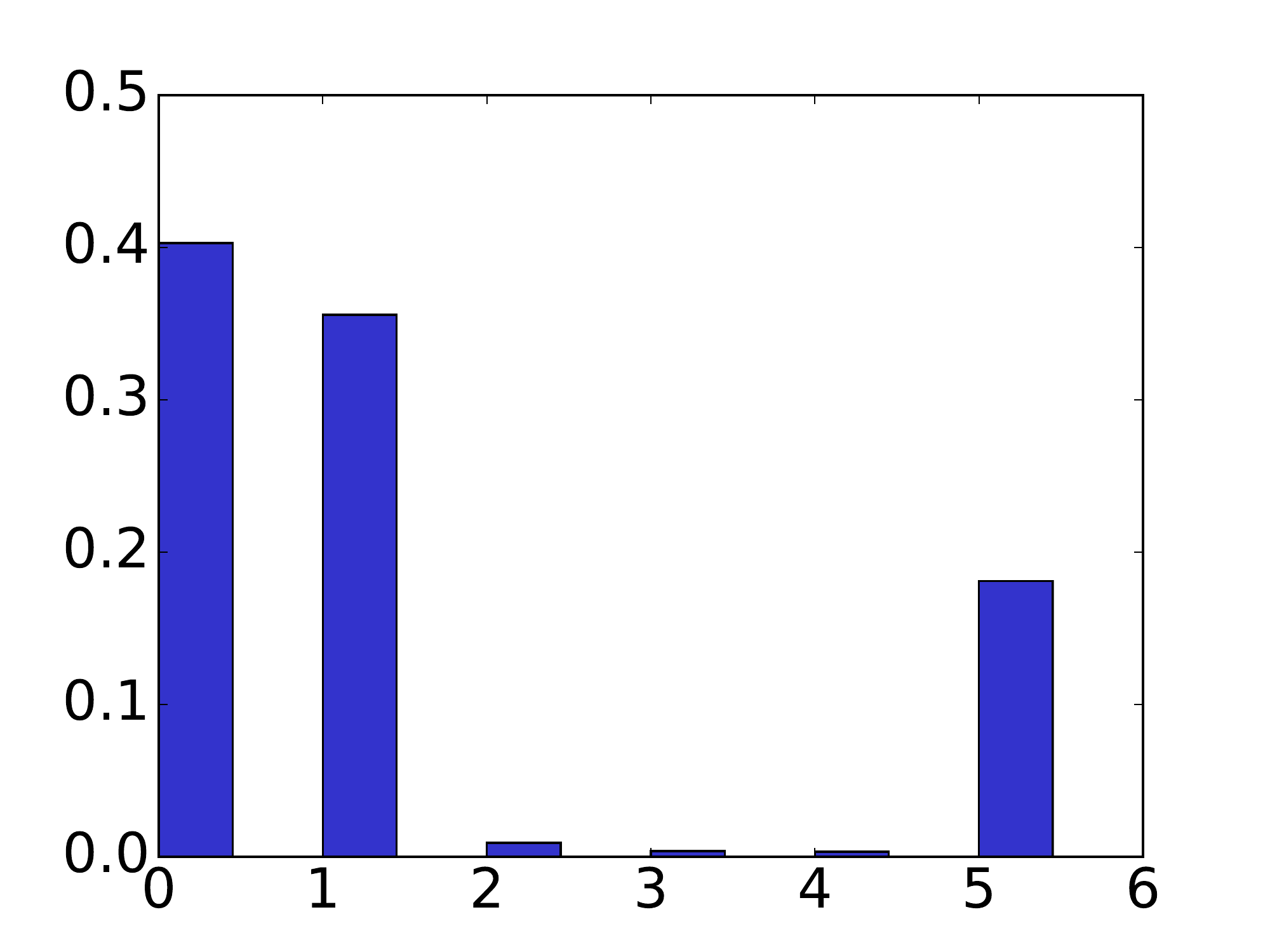}}}
\end{overpic}
}
\subfigure[Original]{
\begin{overpic}[width=0.23\linewidth]{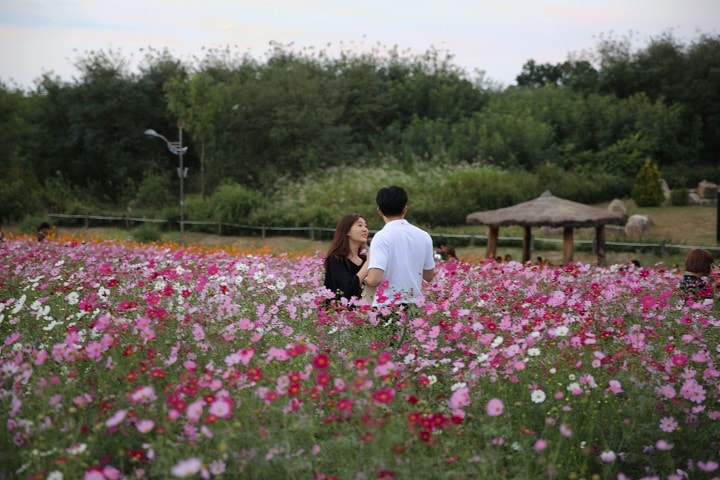}
\put(54.2, 1.8){\fbox{\includegraphics[trim=200px 0px 360px 350px, clip, scale=0.3]{./figure/discussion/L_B_orig}}}
\end{overpic}
}
\subfigure[The manipulated output]
{
\begin{overpic}[width=0.23\linewidth]{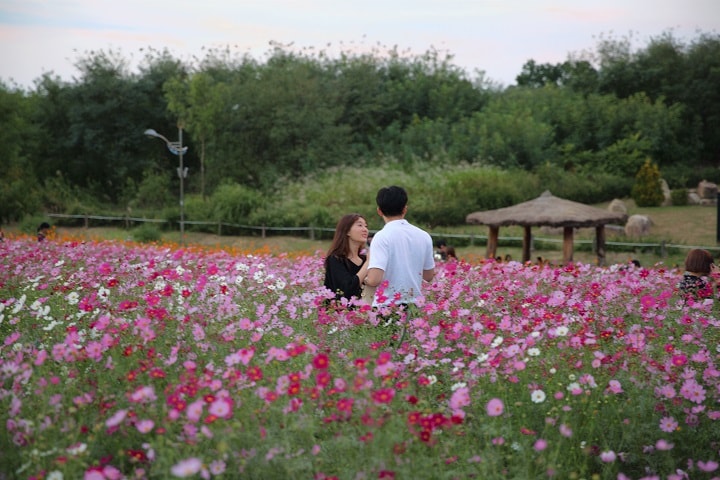}
\put(54.2, 1.8){\fbox{\includegraphics[trim=200px 0px 360px 350px, clip, scale=0.3]{./figure/discussion/L_B_out}}}
\end{overpic}
}
\subfigure[The histogram of the output]{\includegraphics[width=0.23\linewidth]{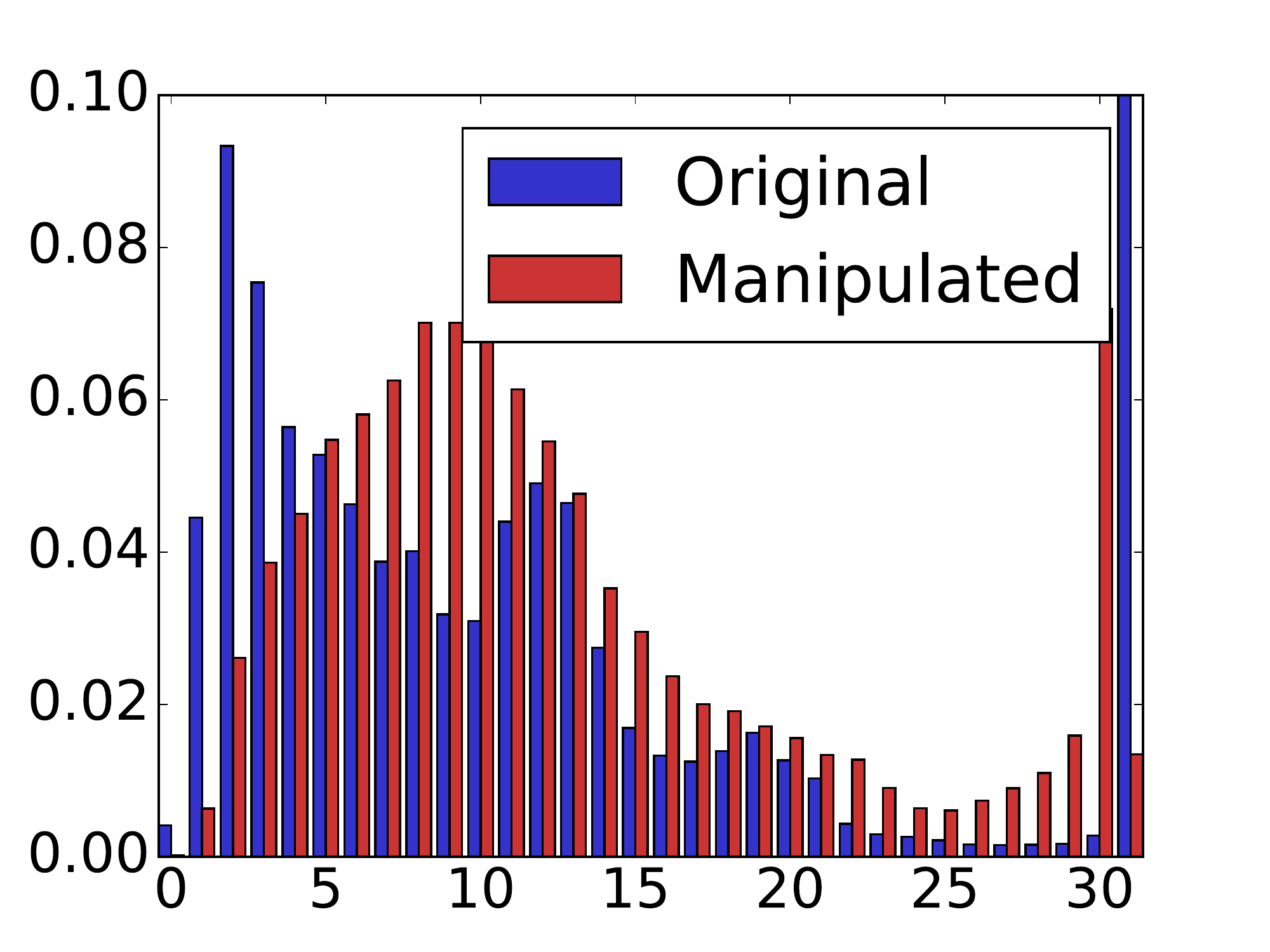}}
\caption{The result of global luminance histogram manipulation. We replace the global luminance histogram of image (b) with that of (a) during the forward process to analyze our network.  (c) and (d) show the result and the change of the histogram. As the deep network recognizes the content of (a) that consists of many dark and bright pixels, we can see that the histogram (red) shifts to the middle from the original (blue).}
\label{fig:discussion_1}
\end{figure*}

\begin{figure*}
\center
\subfigure[An external image]{\includegraphics[width=0.23\linewidth]{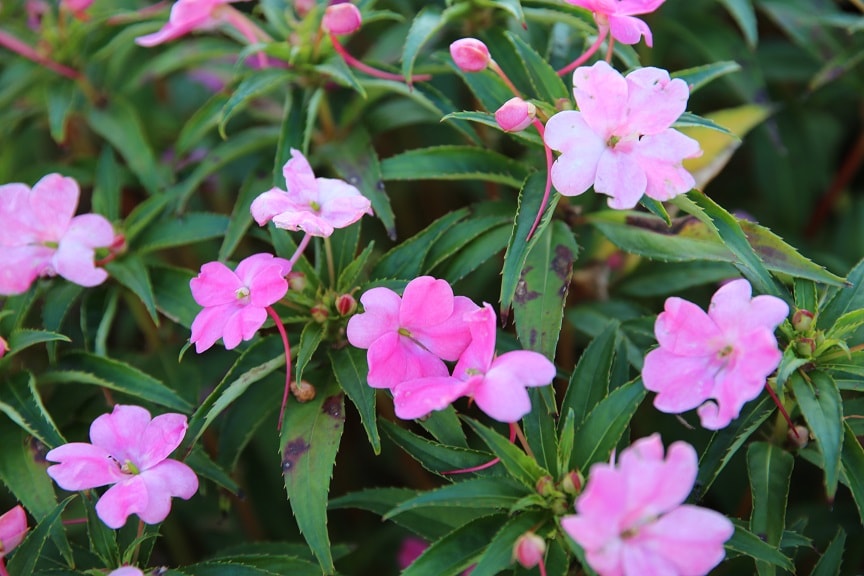}}
\subfigure[Original]{
\begin{overpic}[width=0.23\linewidth]{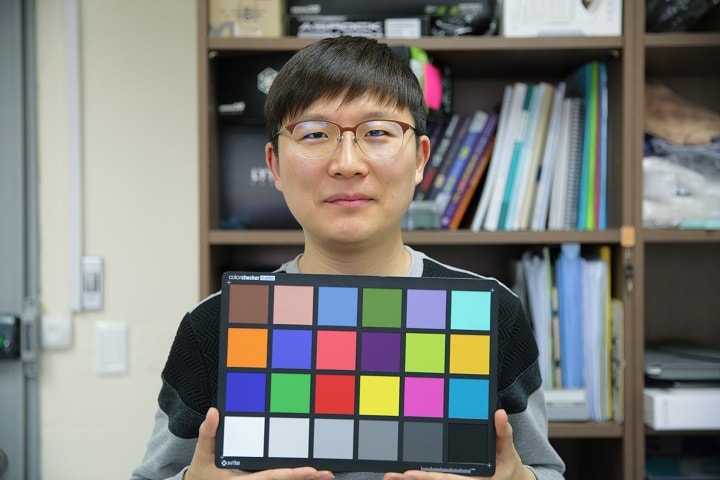}
\put(54.2, 1.8){\fbox{\includegraphics[trim=200px 70px 360px 290px, clip, scale=0.3]{./figure/discussion/rg_B_orig}}}
\end{overpic}
}
\subfigure[The manipulated output]
{
\begin{overpic}[width=0.23\linewidth]{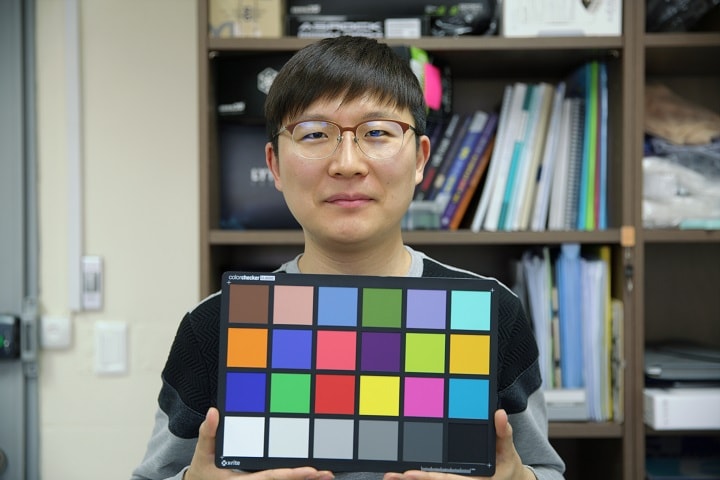}
\put(54.2, 1.8){\fbox{\includegraphics[trim=200px 70px 360px 290px, clip, scale=0.3]{./figure/discussion/rg_B_out}}}
\end{overpic}
}
\subfigure[Errormap]{\includegraphics[width=0.23\linewidth]{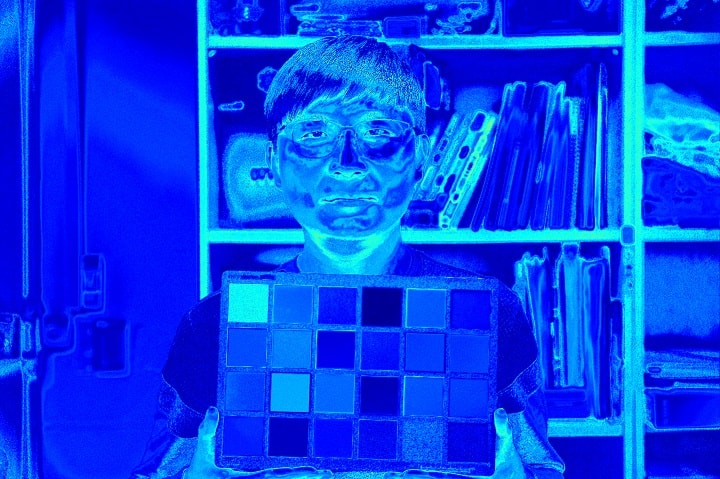}}
\caption{The result of global chromiance histogram manipulation. This time we manipulated the color histogram where (a) is the image from which we extract the global chrominance histogram. (c) shows the manipulated output and (d) shows the errormap between (b) and (c). As the network recognizes the color distribution of image (a), the network modifies color more in the green and the brown regions.}
\label{fig:discussion_2}
\end{figure*}

\subsection{Experimental results}
Table~\ref{table:quantitative} shows the quantitative results using the 50 test images in our dataset for each camera.
In the table, PSNR values comparing the RGB values of the recovered image with the ground truth are reported. 
For both the forward rendering (RAW-to-sRGB) and the reverse rendering (sRGB-to-RAW), the proposed method outperforms the other baseline methods in all categories except for very few Min and MaX errors among test images. 

Results using the MLP were usually worse than the other methods and this indirectly indicates the scene dependency in photographs. 
While the SRCNN showed some ability to deal with the scene dependency, its receptive field is limited to local neighborhoods and it cannot model the global scene context.
The MLP and the SRCNN are optmized to model the mean of the color mapping in dataset and some of high values of the Min and the Max values in the results can be explained that some test images exist around the mean of our dataset.

One can expect that hierarchical CNN features are able to capture the local and the global scene context that are useful for the scene dependent imaging. 
However, the experimental results show that they are not as efficient as our color histogram features.
We attribute the bad performance of the FCN to the fact that the FCN is not sufficiently trained on only hundreds of training examples.
Although we could sufficiently train the HCN through the in-network sampling method~\cite{BansalChen16,Larsson16}, concatenating multi-level upsampled feature maps consume large memory for high-definition images from consumer cameras, which cannot be handled in test time.
In summary, the results clearly show that our deep network that learns the local and the global color distribution is more efficient for accurately modelling the scene dependent image processing in cameras.

\Fref{fig:qualitative} shows some of the examples of the image recovery. 
The figure shows that the proposed method can model the in-camera imaging process accurately in a qualitative way.
It also shows that the other baseline methods also do a reasonable job of recovering images as the scene dependency applies to a set of specific colors or regions.

\begin{figure*}
\subfigure[]{
\begin{tabular}{c@{}c@{}c}
\multicolumn{3}{c}{\includegraphics[width=0.3\linewidth]{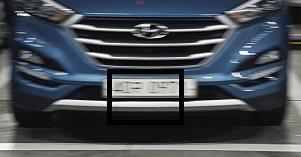}} \\
\includegraphics[width=0.097\linewidth]{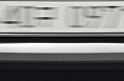}
\includegraphics[width=0.097\linewidth]{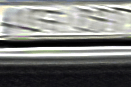}
\includegraphics[width=0.097\linewidth]{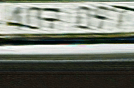}
\end{tabular}
}
\subfigure[]{
\begin{tabular}{c@{}c@{}c}
\multicolumn{3}{c}{\includegraphics[width=0.3\linewidth]{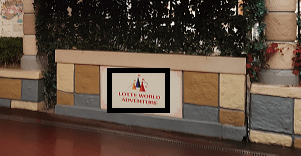}} \\
\includegraphics[width=0.097\linewidth]{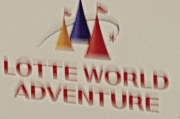}
\includegraphics[width=0.097\linewidth]{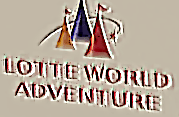}
\includegraphics[width=0.097\linewidth]{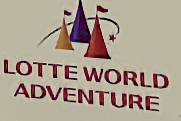}
\end{tabular}
}
\subfigure[]{
\begin{tabular}{c@{}c@{}c}
\multicolumn{3}{c}{\includegraphics[width=0.3\linewidth]{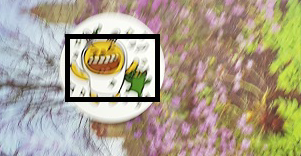}} \\
\includegraphics[width=0.097\linewidth]{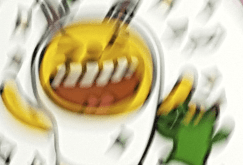}
\includegraphics[width=0.097\linewidth]{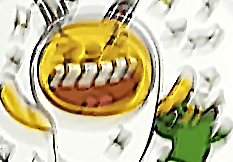}
\includegraphics[width=0.097\linewidth]{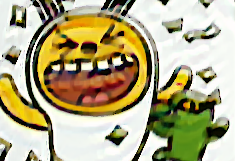}
\end{tabular}
}
\caption{Image deblurring results. Blurred images are shown on top. In the bottom: original blurred image patch, deblurred result using~\cite{pan2016blind} in sRGB space, and deblurred result in RAW space. Deblurring in RAW space outputs much sharper images.}
\label{fig:deblurring}
\end{figure*}

\subsection{Analysis}
We conducted more experiments to analyze the scene dependent processing learned by our network. 
For the analysis, we use two RAW images A and B. 
We first extract the learnable histogram feature from A, and replace the extracted histogram of B with that of A before injecting it to the DNN forward process. 
Note that the RAW image itself is the same, we just simulate the scene context change by changing the histogram. 
The intention of this analysis is to see how our network responds to the changes in the scene context. 

\Fref{fig:discussion_1} shows the result of manipulating the global luminance histogram. 
The histogram in \fref{fig:discussion_1} (a) indicates the luminance distribution of a high contrast image, which is typical for backlit photos. We replace the histogram (a) with that of image (b) during the forward process of (b).
\Fref{fig:discussion_1} (c) and (d) show the result of the manipulation. As can be seen, the deep network brightens shadow regions and darken highlight regions.
The network recognizes many dark regions and bright regions in the given histogram, and compensates by shifting the brightness to the middle (red in \fref{fig:discussion_1} (d)) compared to the original histogram (blue in \fref{fig:discussion_1} (d)). Note that it is what the Auto Lighting Optimizer of Canon cameras does as described in~\cite{CanonBrochure}.

In \fref{fig:discussion_2}, we show the result of manipulating the global chrominance histogram by going through the same process as explained above. 
By changing the global color context of image B with that of A, we can observe that our network responses more strongly to green and brown colors than the original.
We can interpret this as our network recognizing the context of A as a natural scene from color distribution, trying to make natural objects like trees more visually pleasing. 

These examples show that our deep network recognizes specific scene context such as high contrast or nature images, and manipulates the brightness and colors to make images more visually pleasing as done in the scene dependent imaging pipeline of cameras. 
The experiments  also show that the deep network does not memorize each example and can infer the mapping under various scene contexts.

\subsection{Application to Image Deblurring}
To show the effectiveness of the proposed method, we apply it to the image deblurring application.
It is well known that the blur process actually happens in the RAW space, but most deblurring algorithms are applied to the sRGB images since the RAW images are usually unavailable.
Tai \etal~\cite{Tai:2013} brought up this issue and showed that being able to linearize the images have a significant effect in the deblurred results. 
However, the radiometric calibration process in \cite{Tai:2013} is rather limited and can only work under manual camera settings. 

We show that we can improve the deblurring performance on images taken from a smartphone camera (Samsung Galaxy S7) in automode. 
To do this, we use the image deblurring method of Pan \etal~\cite{pan2016blind}, which is a blind image deblurring method that uses the dark channel prior. We use the source code from the authors’ website and the default settings except for the kernel size.
The RAW images are first computed from the corresponding sRGB images using the sRGB-to-RAW rendering of the proposed method, deblurred, and converted back to sRGB images using the RAW-to-sRGB rendering of the proposed method.

\Fref{fig:deblurring} shows the deblurring results. 
As expected, the deblurring method~\cite{pan2016blind} does not work well on nonlinear sRGB images and 
there are some artifacts on deblurred scenes.
On the other hand, the deblurring algorithm works well using our framework.
The recovered images are much sharper and there are no significant artifacts.

\section{Conclusion}
In this paper, we presented a novel deep neural network architecture that can model the scene dependent image processing inside cameras. 
Compared to previous works that employ imaging models that are scene independent and can only work for images taken under the manual mode, 
our framework can be applied to the images that are taken under the auto-mode including photos from smartphone cameras.
We also showed the potential of applying the proposed method for various computer vision tasks via image deblurring examples.

\section*{Acknowledgement}
This work was supported by Global Ph.D. Fellowship Program through the National Research Foundation of Korea (NRF) funded by the Ministry of Education (NRF-2015H1A2A1033924), and the National Research Foundation of Korea (NRF) grant funded by the Korea government (MSIP) (NRF-2016R1A2B4014610).

{\small
\bibliographystyle{ieee}
\bibliography{egbib}
}

\end{document}